\documentclass[letterpaper,10pt,conference]{ieeeconf}  % Comment this line out if you need a4paper

\IEEEoverridecommandlockouts                              % This command is only needed if 
\usepackage{graphics,graphicx} % for pdf, bitmapped graphics files
\usepackage{epsfig} % for postscript graphics files
\usepackage{amsmath} % assumes amsmath package installed
\usepackage{amssymb}  % assumes amsmath package installed
\usepackage{enumerate}
\usepackage{flushend}
\usepackage{cite}
\usepackage{color}
\usepackage{float}
\usepackage{wasysym}
\def\bxone{\mbox{${\hspace*{-0.2mm}}^1$\hspace*{-0.4mm}\boldmath$x$}}
\def\bxtwo{\mbox{${\hspace*{-0.2mm}}^2$\hspace*{-0.3mm}\boldmath$x$}}
\def\Tone{\mbox{${\hspace*{-0.3mm}}^{1}$\hspace*{-0.3mm}$T$}}
\def\Ttwo{\mbox{${\hspace*{-0.3mm}}^{2}$\hspace*{-0.2mm}$T$}}
\def\Tones{\mbox{\scriptsize{${\hspace*{-0.3mm}}^{1}$\hspace*{-0.3mm}$T$}}}
\def\Ttwos{\mbox{\scriptsize{${\hspace*{-0.3mm}}^{2}$\hspace*{-0.3mm}$T$}}}
\def\bEta{\mbox{\boldmath$\eta$}}
\def\bAs{\mbox{\scriptsize{\boldmath$A$}}}
\def\rme{\mbox{${\rm e}$}}
\def\bth{\mbox{\boldmath$\theta$}}
\def\ddth{\mbox{$\ddot{\theta}$}}

\def\dth{\mbox{$\dot{\theta}$}}

\def\bR{\mbox{\boldmath$R$}}
\def\bp{\mbox{\boldmath$p$}}

\def\bb{\mbox{\boldmath$b$}}

\def\bg{\mbox{\boldmath$g$}}

\def\bM{\mbox{\boldmath$M$}}

\def\bb{\mbox{\boldmath$b$}}

\def\bA{\mbox{\boldmath$A$}}
\def\bI{\mbox{\boldmath$I$}}
\def\bG{\mbox{\boldmath$G$}}

\def\bx{\mbox{\boldmath$x$}}

\def\bX{\mbox{\boldmath$X$}}

\def\bS{\mbox{\boldmath$S$}}

\def\bQ{\mbox{\boldmath$Q$}}

\title{\LARGE \bf
Stability Principle Inherent in Wheel Gait of Planar X-shaped Walker Generated Using Constant Torque Drive and Mechanical Stoppers}

\author{Fumihiko Asano$^{1}$ and Cong Yan$^{2}$
\thanks{This research was partially supported by Grant-in-Aid for Scientific Research (C) No. 23K03727, provided by the Japan Society for the Promotion of Science (JSPS).}
\thanks{$^{1}$F. Asano is with the Graduate School of Advanced Science and Technology, Japan Advanced Institute of Science and Technology, 1-1 Asahidai, Nomi, Ishikawa 923-1292, Japan {\tt\small fasano@jaist.ac.jp}}
\thanks{$^{2}$C. Yan is with the Department of Mechanical Engineering, Ritsumeikan University, 1-1-1 Nojihigashi, Kusatsu, Shiga 525-8577, Japan {\tt\small gansou@fc.ritsumei.ac.jp}}%
}

\begin{document}

\maketitle
\thispagestyle{empty}
\pagestyle{empty}

\setlength{\arraycolsep}{1.3pt}
%\setlength{\abovedisplayskip}{5.8pt}
%\setlength{\belowdisplayskip}{5.8pt}
%\renewcommand{\arraystretch}{1.0}
%%%%%%%%%%%%%%%%%%%%%%%%%%%%%%%%%%%%%%%%%%%%%%%%%%%%%%%%%%%%%%%%%%%%%%%%%%%%%%%%
\begin{abstract}
Since the late 19th century when the first walking toys were developed, it has been known that mechanical stoppers of the hip joint is crucial for generating stable passive dynamic walking. Recent research on passive-dynamic and limit-cycle walkers has also confirmed that mechanical stoppers at the hip and knee joints are effective for generating stable walking motion, but theoretical research on how this mechanical constraint enhances the overall gait stability remains insufficient. This paper introduces a planar X-shaped walker equipped with mechanical stoppers at the hip joint, and investigates the effect of the mechanical constraint on the stability of the wheel gait generated by constant torque drive on a downslope. By simply falling forward while using the stoppers to constrain itself to the target impact posture, the robot can generate a highly stable wheeled gait. We divide the motion of one step into four phases, derive approximate analytical solutions for the state error transition function matrix in each phase using a linearized model, and analyze the increase or decrease in the state error norm using metrics such as its maximum singular value. Numerical simulations demonstrate that while two phases are unstable in terms of the increase in the state error norm, the remaining two phases are stable.
%Specifically, using a linearized model, we divide the motion of one step into four phases, and theoretically and numerically analyze the characteristics of state error transitions in each phase. 

\end{abstract}
%%%%%%%%%%%%%%%%%%%%%%%%%%%%%%%%%%%%%%%%%%%%%%%%%%%%%%%%%%%%%%%%%%%%%%%%%%%%%%%%
\section{INTRODUCTION}

Through his pioneering research on passive dynamic walking, McGeer demonstrated that the passive compass-like biped robot can generate stable walking motion on downslopes by repeatedly performing the natural swinging motion of the legs based on gravity and the collision of stance-leg exchange \cite{McGeer}. Simultaneously, he showed that passive-dynamic walkers with knee joints require mechanical stoppers to be attached to both knees for stable motion generation \cite{McGeer2}. Walking toys also possess stoppers somewhere on their bodies and utilize this mechanical constraint to walk down slopes with a steady, rhythmic gait \cite{Fallis,Willson}. Through these pioneering studies and toy development, insights emerged that using mechanical stoppers to maintain impact postures as consistently as possible is effective for generating continuous walking motion and enhancing the stability of hybrid zero dynamics \cite{Grizzle,HZD}.

The rimless wheel (RW) is also a commonly used model for analyzing passive dynamic walking \cite{Coleman,Coleman2}. It possesses a distinct, self-evident stability compared to the aforementioned passive-dynamic walkers for the following reasons. The RW with a symmetrical shape occurs each collision in the same posture, resulting in a constant ratio of kinetic energy immediately before and immediately after collision. Simultaneously, the restored mechanical energy per step due to gravity also remains constant. Consequently, the generated motion is always period-1 and asymptotically stable. The above can be easily explained using the recurrence formula for kinetic energy immediately before impact, indicating gait stability at an obvious level. In the wheel gait addressed in this paper, however, the inclusion of kinetic energy-loss term that is difficult to describe analytically immediately makes stability analysis extremely challenging. While the convergence of gait descriptors is very similar to that of a RW, the difficulty in providing a mathematical explanation is worlds apart.

This paper examines the stability principles inherent in wheel gaits generated while constrained to target impact postures using mechanical stoppers, employing the following methods. First, we derive the equations of motion, collisions, and constraint conditions for a planar 2-DOF X-shaped walker. We then numerically demonstrate that a stable wheel gait can be generated on a downward slope by causing the mechanical stopper on the swing leg to collide with the stance leg, causing the robot to fall forward. Next, using a linearized model approximating the equation of motion around an unstable equilibrium point, we derive the state error transition function matrices for the two swing phases and the two collision phases. For the two collision phases, we derive the matrices directly using the collision equations from the nonlinear model. By calculating the eigenvalues and maximum singular values of these matrices, we can gain some insight into the trend of increase or decrease in the state error norm during each phase. Furthermore, by sequentially multiplying the four matrices to reduce dimensions, we can derive the state error transition function from the angular velocity error of the entire body immediately before impact to the next. Through numerical simulations, we clarify that walking speed and convergence characteristics change depending on the magnitude of the constant torque applied to the hip joint, and that while two of the four phases are unstable, the remaining two are stable.
\section{MATHEMATICAL MODELING AND WHEEL GAIT GENERATION}

\subsection{Model Assumptions}

\begin{figure}[!t]
\centering
\vspace*{2mm}
\includegraphics[width=0.98\linewidth]{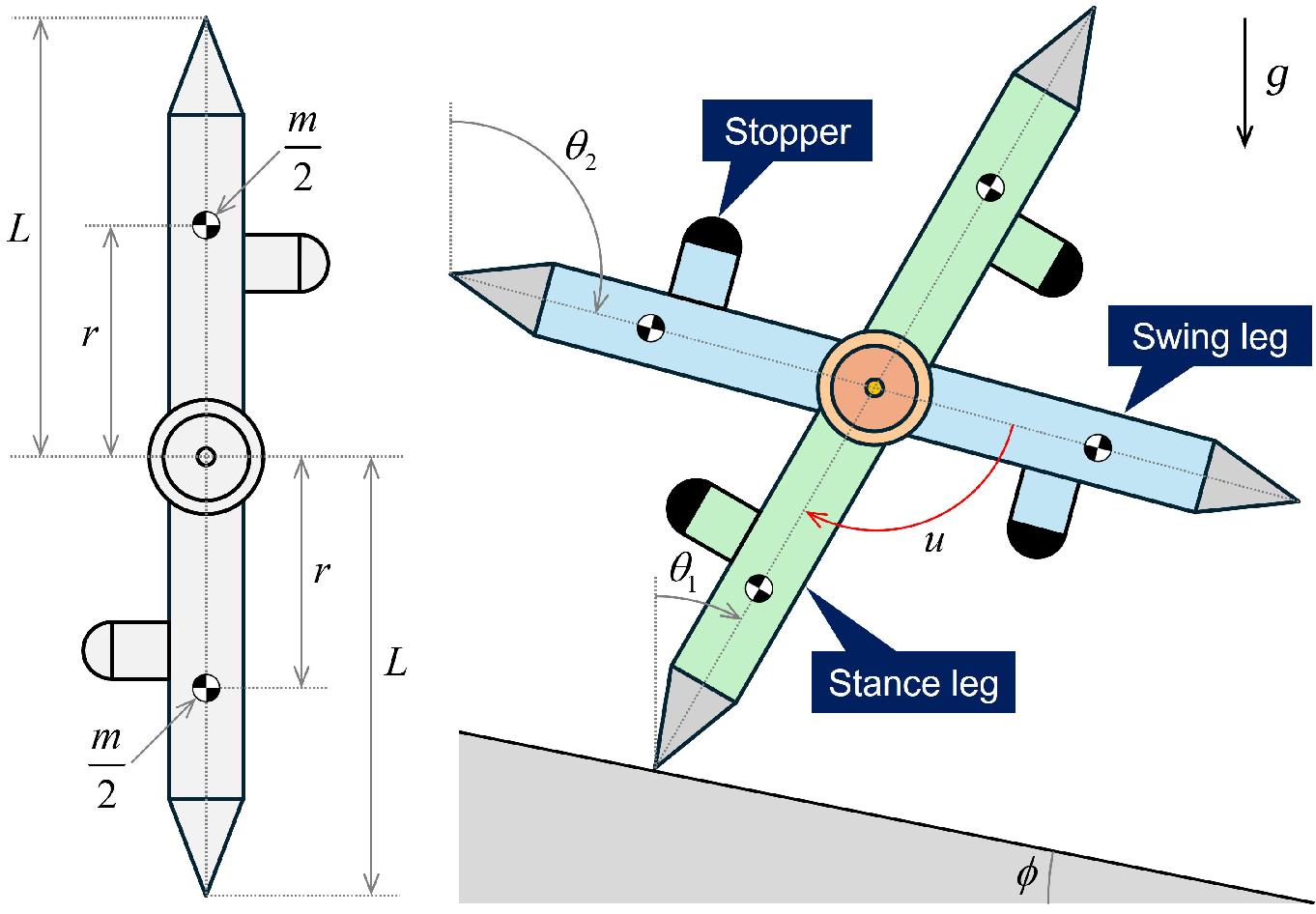}\\
\vspace*{-2mm}
\caption{Model of planar X-shaped walker with mechanical stoppers}
\label{fig2.01}
\vspace*{-3mm}
\end{figure}

Figure \ref{fig2.01} shows the model of the planar X-shaped walker with stoppers walking on a downhill slope. Two identical leg frames are connected to each other via coaxial active joints, and $L$ is half the length of the leg frame. The mass of each leg frame is $m$, with half of this mass located at a distance $r$ from the hip joint. This results in a moment of inertia $I = mr^2$ about the center of mass (COM) or hip joint. $\theta_1$ and $\theta_2$ are the absolute angles of the stance and swing legs relative to the vertical up direction. The motor attached to the hip joint can apply a control torque $u$ with a positive clockwise direction from the swing leg to the stance leg. We assume that the stance foot is in contact with the ground at one point without sliding. 

The most distinctive feature of the bipedal robot discussed in this paper is the installation of mechanical stoppers on each leg frame. When the stoppers on the swing leg undergo a fully inelastic collision with the stance leg, the robot falls forward while maintaining the smaller hip-joint angle at $\alpha$. Since the mass of the stoppers is assumed negligible, however, the equation of motion becomes very simple and almost linear.

We consider the motion from immediately before the collision of the stance-leg exchange to the next one as one-step motion; this consists of two collision phases and two swing phases. Figure \ref{fig2.02} shows the phase sequence in one-step walking motion. We will describe the mathematical model for each phase sequentially, setting the generalized coordinate vector $\bth = \left[
\begin{array}{cc}
\theta_1 & \theta_2
\end{array}
\right]^{\rm T}$.

\subsection{Collision Phase 1}

\begin{figure}[!t]
\centering
\vspace*{2mm}
\includegraphics[width=0.98\linewidth]{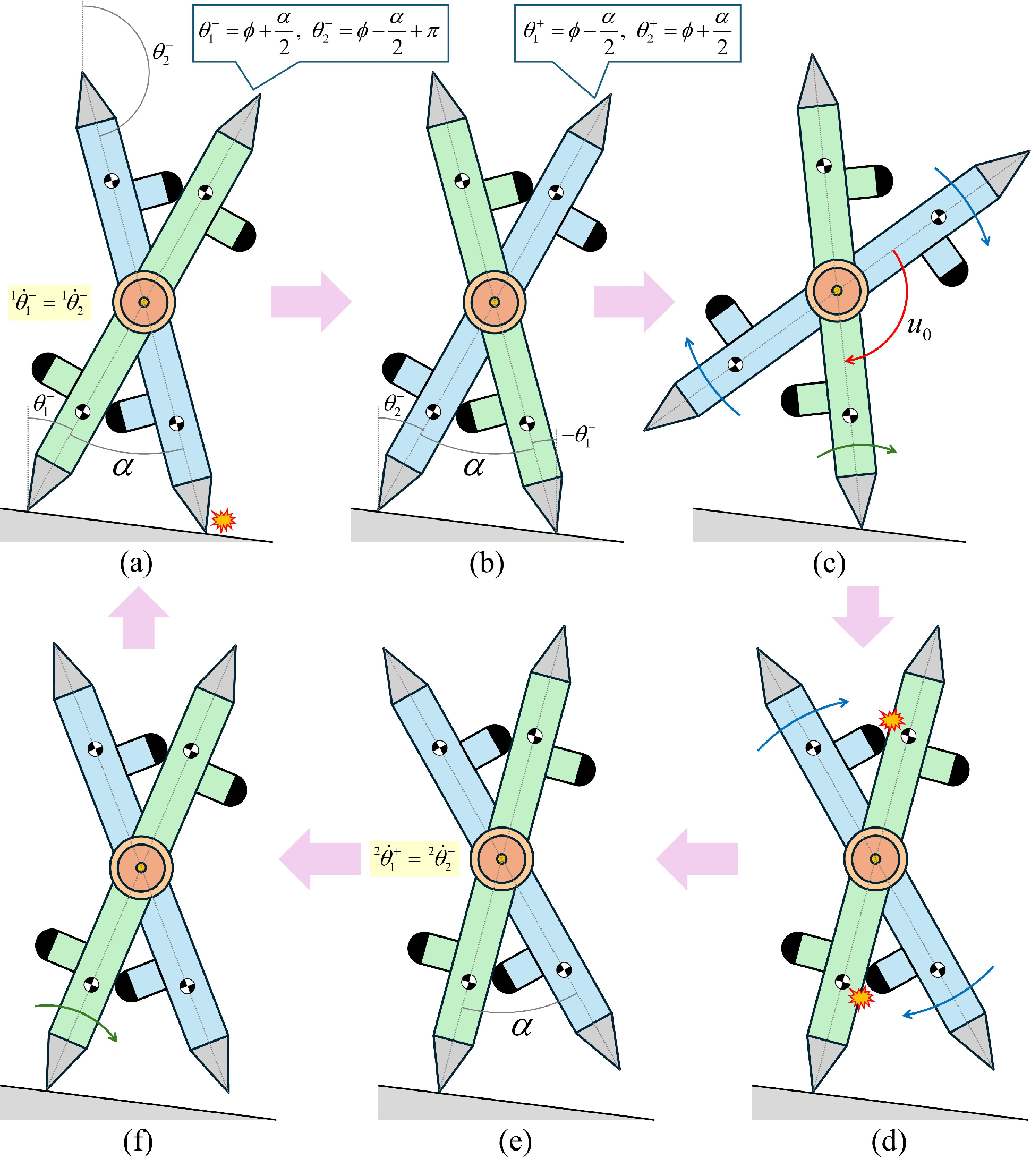}\\
\vspace*{-4mm}
\caption{Phase sequence in one-step walking motion}
\label{fig2.02}
\vspace*{-3mm}
\end{figure}

The collision phase during which the stance leg is exchanged shall be referred to as Collision Phase 1 (CP1). Considering that the angles of the stance leg and swing leg are exchanged, and noting that the angle of the stance leg immediately after impact, as shown in Figs. \ref{fig2.02}(a) and (b), must be obtained by subtracting $\pi$ [rad] from the angle of the swing leg immediately before impact, the update rule for the angular position vector can be summarized as follows.
\begin{equation}
\bth^+ = 
\left[
\begin{array}{cc}
0 & 1 \\
1 & 0
\end{array}
\right]
\bth^- - \left[
\begin{array}{c}
\pi \\
0
\end{array}
\right] = \left[
\begin{array}{c}
\phi - \frac{\alpha}{2} \\
\phi + \frac{\alpha}{2}
\end{array}
\right]
\label{eq2.01}
\end{equation}
The details of the derivation process for the angular velocity update rule are omitted. Considering the leg exchange, it can ultimately be summarized by the following equation.
\begin{equation}
\dot{\bth}^+ = 
\left[
\begin{array}{cc}
\frac{2mL^2 \cos \alpha}{2mL^2+I} & \frac{I}{2mL^2+I} \\
1 & 0
\end{array}
\right]
\dot{\bth}^-
\label{eq2.02}
\end{equation}
In the wheel gait considered in this paper, it is assumed that $\dth_1^- = \dth_2^-$ immediately before each swing-leg landing due to the mechanical constraint of the stopper. Therefore, Eq. (\ref{eq2.02}) can be further simplified as follows.
\begin{equation}
\dot{\bth}^+ = 
\left[
\begin{array}{c}
\frac{2mL^2 \cos \alpha + I}{2mL^2+I} \\
1 
\end{array}
\right]
\dth_1^-
\label{eq2.03}
\end{equation}

\subsection{Swing Phase 1}

The robot equation of motion becomes
\begin{equation}
\left[
\begin{array}{cc}
2 m L^2 + I & 0 \\
0 & I
\end{array}
\right]
\left[
\begin{array}{c}
\ddth_1 \\
\ddth_2
\end{array}
\right] + 
\left[
\begin{array}{c}
-2mgL \sin \theta_1 \\
0
\end{array}
\right] = 
\left[
\begin{array}{c}
1 \\
-1
\end{array}
\right] u ,
\label{eq2.04}
\end{equation}
and hereafter, this is denoted as
\begin{equation}
\bM \ddot{\bth} + \bg = \bS u.
\label{eq2.05}
\end{equation}
Eq. (\ref{eq2.04}) clearly shows that the equation of motion is almost linear, meaning the inertia matrix is a constant matrix, there are no nonlinear velocity terms, and only the gravitational term contains a simple nonlinear term \cite{Kiefer,Spong,Agrawal}. The only nonlinear term within the gravitational term is $\sin \theta_1$. This paper employs a linearized model obtained by approximating this term linearly around the unstable equilibrium point $\theta_1 = 0$ for analysis.

As mentioned, a constant negative torque $u = u_0 \ (<0)$ is continuously applied to the hip joint during this phase. Consequently, the swing leg is driven clockwise, while the stance leg is driven counterclockwise due to the reaction torque. On a downward slope, however, the stance leg can utilize gravitational force, enabling it to continue moving clockwise at a slower speed than the swing leg. We define this phase, where both legs move clockwise, as Swing Phase 1 (SP1).

\subsection{Collision Phase 2}

As shown in Figs. \ref{fig2.02}(d) and (e), we assume that the stoppers attached to the swing leg make a perfectly inelastic collision with the stance leg. Immediately after this collision, the hip joint angle is fixed at $\bS^{\rm T} \bth = \alpha - \pi $. This collision phase shall be referred to as Collision Phase 2, and the equation for this perfectly inelastic collision is specified as
\begin{equation}
\bM \dot{\bth}^+ = \bM \dot{\bth}^- + \bS \lambda_I,
\label{eq2.06}
\end{equation}
where $\lambda_I$ is the impulse. The velocity constraint condition that should be satisfied immediately after impact is written as
\begin{equation}
\bS^{\rm T} \dot{\bth}^+ = 0.
\label{eq2.07}
\end{equation}
By solving Eqs. (\ref{eq2.06}) and (\ref{eq2.07}), $\lambda_I$ is obtained as
\begin{equation}
\lambda_I = -\left(
\bS^{\rm T} \bM^{-1} \bS
\right)^{-1} \bS^{\rm T} \dot{\bth}^-.
\label{eq2.08}
\end{equation}
By substituting this into Eq. (\ref{eq2.06}) and arranging, the angular velocity vector immediately after impact is obtained as
\begin{equation}
\dot{\bth}^+ =  \left( \bI_2 - \bM^{-1} \bS \left(
\bS^{\rm T} \bM^{-1} \bS
\right)^{-1} \bS^{\rm T} \right) \dot{\bth}^-.
\label{eq2.09}
\end{equation}

\subsection{Swing Phase 2}

After CP2, assuming the swing leg's stoppers remain in contact with the stance leg without separating, the robot will fall forward as a 1-DOF rigid body. This can also be interpreted as a constant torque being applied to the hip joint, forcing the swing leg against the stance leg. We will refer to this phase as Swing Phase 2 (SP2). The velocity constraint condition during this phase is written as
\begin{equation}
\bS^{\rm T} \dot{\bth} = 0,
\label{eq2.10}
\end{equation}
and the robot equation of motion with constraint torque $\lambda$ then becomes
\begin{equation}
\bM \ddot{\bth} + \bg = \bS \lambda.
\label{eq2.11}
\end{equation}
The time derivative of Eq. (\ref{eq2.10}) becomes
\begin{equation}
\bS^{\rm T} \ddot{\bth} = 0 = \bS^{\rm T} \bM^{-1} \left( \bS \lambda - \bg \right),
\label{eq2.12}
\end{equation}
and we can solve this for $\lambda$ as
\begin{equation}
\lambda = \left(
\bS^{\rm T} \bM^{-1} \bS
\right)^{-1} \bS^{\rm T} \bM^{-1} \bg.
\label{eq2.13}
\end{equation}
By substituting this into Eq. (\ref{eq2.11}) and arranging, the equation of motion is finally arranged as
\begin{equation}
\bM \ddot{\bth} = 
\left( \bS
\left(
\bS^{\rm T} \bM^{-1} \bS
\right)^{-1} \bS^{\rm T} \bM^{-1} - \bI_2
\right)
\bg.
\label{eq2.14}
\end{equation}
The robot then falls forward as a 1-DOF rigid body.

\begin{figure}[!b]
\vspace*{-2mm}
\footnotesize
\centering
\vspace*{0.5mm}
\hspace*{0.5mm}
\scalebox{0.66}{\input{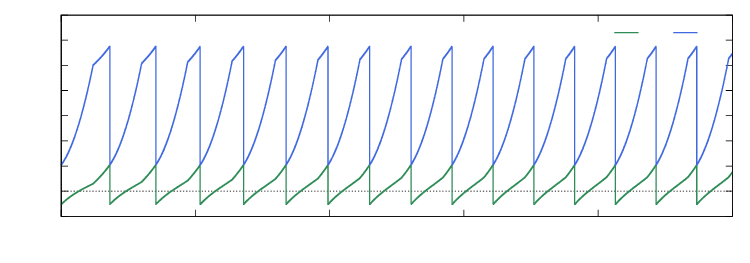}}\\
\vspace*{-1mm}
(a) Angular positions \\
\hspace*{0.5mm}
\scalebox{0.66}{\input{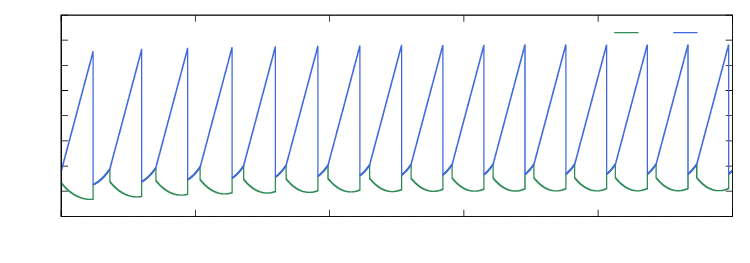}}\\
\vspace*{-1mm}
(b) Angular velocities \\
\hspace*{0.5mm}
\scalebox{0.66}{\input{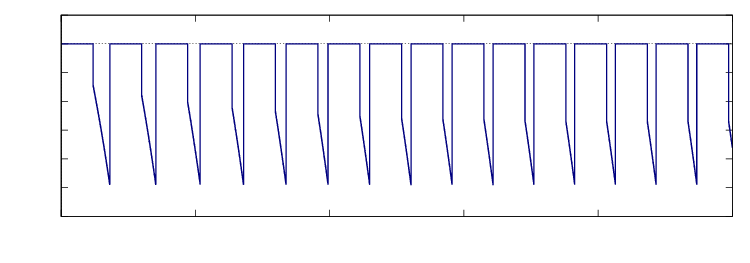}}\\
\vspace*{-1mm}
(c) Constraint torque \\
\hspace*{0.5mm}
\scalebox{0.66}{\input{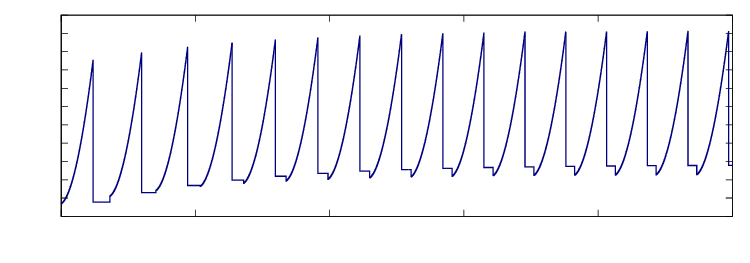}}\\
\vspace*{-1mm}
(d) Total mechanical energy \\
\vspace*{-2mm}
\caption{Simulation results of wheel gait generation}
\label{fig2.03}
\end{figure}

\begin{table}[!b]
\vspace*{-1mm}
\caption{Physical and control parameters}
\label{table01}
\vspace*{-1.5mm}
\centering
\renewcommand{\arraystretch}{1.30}
\begin{tabular}{ccc} \hline
$m$ & 1.0 & kg \\
$L$ & 1.0 & m \\
$r$ & 0.5 & m \\
$I \left( = m r^2 \right)$ & 0.25 & kg$\cdot$m$^2$ \\
\hline
\end{tabular}
\hspace*{1mm}
\begin{tabular}{ccc} \hline
$\alpha$ & $\frac{\pi}{4}$ & rad \\
$\phi$ & 0.13 & rad \\ 
$u_0$ & $-2.5$ & N$\cdot$m \\ 
\hline
\end{tabular}
\end{table}

\subsection{Typical Wheel Gait}

Fig. \ref{fig2.03} shows the simulation results of generating an asymptotically stable wheel gait on a downhill slope. Here, (a) is the angular positions, (b) the angular velocities, (c) the constraint torque, and (d) the total mechanical energy. The physical and control parameters were chosen as the values listed in Table \ref{table01}. The robot started walking from the target impact posture setting the initial angular velocities of Eq. (\ref{eq2.03}) with $\dth_1^- = 0.8$ [rad/s]. From Fig. \ref{fig2.03}(a), we can see that both legs continue rotating in the same direction. From Fig. \ref{fig2.03}(b), we can see that after CP2, the angular velocities of both legs remain consistent, meaning the mechanical constraint imposed by the stopper is maintained. From Fig. \ref{fig2.03}(c), regarding the constraint torque during SP2, the relationship $u_0 < \lambda < 0$ holds, and this indicates that applying the constant torque continuously can maintain the constraint condition. From Fig. \ref{fig2.03}(d), we can see that the total mechanical energy monotonically increases due to the constant torque drive during SP1 and decreases significantly during CP2, resulting in a small total restored mechanical energy per step. As the generated motion asymptotically converges, we observe the gradual equilibrium between the amount of energy restored and the amount dissipated. We will revisit this property in Section IV.

\section{DERIVATION OF STATE ERROR TRANSITION FUNCTION MATRICES}

\subsection{Assumptions and Collision Phase 1}

Hereinafter, let $i$ be a nonnegative integer representing the number of steps. We assume that the robot starts walking motion from the target impact posture immediately after impact; this is defined as the 0-th impact for stance-leg exchange. 

The robot is assumed to initiate motion from the target impact posture immediately the collision for stance-leg exchange, defining this as 0-th CP1. The next CP1 is defined as the 1-st CP1, and the motion between immediately before the 0-th and immediately before the 1-st CP1s is called the ``0-th step'' motion. The subsequent CP1s and steps are contextually counted.

%The impact from the $i$-th impact to the $i+1$-th impact is defined as the $i$-th step period, $T_i$. The steady step period is then written as $T_{\infty}$.

Let ${}^{1}\hspace*{-0.3mm}\bx_i^{\pm}$ be the state vectors immediately before and immediately after the $i$-th CP1, ${}^{1}\hspace*{-0.3mm}\bx_{\infty}^{\pm}$ be those in the steady gait, and $\Delta {}^{1}\hspace*{-0.3mm}\bx_i^{\pm}$ be the error of the former with respect to the latter, that is, ${}^{1}\hspace*{-0.3mm}\bx_i^{\pm} = {}^{1}\hspace*{-0.3mm}\bx_{\infty}^{\pm} + \Delta {}^{1}\hspace*{-0.3mm}\bx_{i}^{\pm}$.

From Eqs. (\ref{eq2.01}) and (\ref{eq2.02}), the state transition from immediately before to immediately after the $i$-th CP1 is described as
\begin{equation}
\bxone_i^+ \! = \! \bR_1 \bxone_i^- - 
\left[
\begin{array}{c}
\pi \\
{\bf 0}_{3 \times 1}
\end{array}
\right] , \ 
\bR_1 \! := \! \left[
\begin{array}{cccc}
0 & 1 & 0 & 0 \\
1 & 0 & 0 & 0 \\
0 & 0 & \frac{2mL^2 \cos \alpha}{2mL^2+I} & \frac{I}{2mL^2+I} \\
0 & 0 & 1 & 0
\end{array}
\right],
\label{eq3.01}
\end{equation}
whereas in the steady-state motion it is described as
\begin{equation}
\bxone_{\infty}^+ = \bR_1 \bxone_{\infty}^- - 
\left[
\begin{array}{c}
\pi \\
{\bf 0}_{3 \times 1}
\end{array}
\right].
\label{eq3.02}
\end{equation}
Subtracting Eq. (\ref{eq3.02}) from Eq. (\ref{eq3.01}), we obtain $\Delta \bxone_i^+ = \bR_1 \Delta \bxone_i^-$, which indicates that the state error transition function matrix for the collision phase is $\bR_1$. Since the four eigenvalues of $\bR_1$ are $1$, $-1$, and the two stable modes, CP1 is a stable phase.

\subsection{Swing Phase 1}

Regarding the gravity term in Eq. (\ref{eq2.04}), we consider the following approximate linearization around an unstable equilibrium point.
\begin{equation}
\left[
\begin{array}{c}
-2mgL \sin \theta_1 \\
0
\end{array}
\right] 
 \approx 
\left[
\begin{array}{cc}
-2mgL & 0 \\
0 & 0
\end{array}
\right] 
\left[
\begin{array}{c}
\theta_1 \\
\theta_2
\end{array}
\right] =: \bG_0 \bth
\label{eq3.03}
\end{equation}
The linearized equation of motion becomes $\bM \ddot{\bth} + \bG_0 \bth = \bS u_0$, and its state-space realization becomes
\begin{equation}
\dot{\bx} = \bA_1 \bx + \bb_1,
\label{eq3.04}
\end{equation}
where
\begin{eqnarray*}
\bA_1 &=& 
\left[
\begin{array}{cc}
{\bf 0}_{2 \times 2} & \bI_2 \\
-\bM^{-1} \bG_0 & {\bf 0}_{2 \times 2}
\end{array}
\right] = 
\left[
\begin{array}{cccc}
0 & 0 & 1 & 0 \\
0 & 0 & 0 & 1 \\
\frac{2 m g L}{2 mL^2+I} & 0 & 0 & 0 \\
%\omega_1^2 & 0 & 0 & 0 \\
0 & 0 & 0 & 0
\end{array}
\right], \\
% \ \omega_1 = \sqrt{\frac{2 m g L}{2 mL^2+I}} \\
\bb_1 &=& 
\left[
\begin{array}{c}
{\bf 0}_{2 \times 1} \\
\bM^{-1} \bS u_0
\end{array}
\right] = 
\left[
\begin{array}{cccc}
0 & 0 & 
\frac{u_0}{2mL^2+I}
 & 
-\frac{u_0}{I}
\end{array}
\right]^{\rm T}.
\label{eq3.05}
\end{eqnarray*}

The state error transition function matrix for this phase is derived below. Hereafter, $t$ is defined as a time variable reset to zero at every instant of collision. Immediately after CP1, i.e., at $t=0^+$, the robot initiates single-limb support motion and continues driving with a constant hip-joint torque until the mechanical stoppers attached to the swing leg collides with the stance leg. This collision time is denoted as $t = \Tone_i$. From Eq. (\ref{eq3.04}), the state vector immediately before the $i$-th CP2, $\bxtwo_{i}^-$, is obtained with the state vector immediately after the $i$-th CP1, $\bxone_i^+$, as the initial state as follows.
\begin{eqnarray}
\hspace*{-5mm}
\bxtwo_{i}^- &=& \bx \left( \Tone_i^- \right) = 
{\rm e}^{\bAs_1 \Tones_i} \bxone_i^+ + \int_{0^+}^{\Tones_i^-} {\rm e}^{\bAs_1 \left( \Tones_i - \tau \right)} \bb_1 \ {\rm d} \tau 
\nonumber \\
&=& {\rme}^{\bAs_1 \Tones_i} 
\left(
\bxone_i^+ + \int_{0^+}^{\Tones_i^-} {\rm e}^{-\bAs_1 \tau} \bb_1 \ {\rm d} \tau
\right) 
\nonumber \\
&=& {\rme}^{\bAs_1 \Tones_{\infty}} {\rme}^{\bAs_1 \Delta \Tones_i} \nonumber \\
& & \times
\left(
\bxone_{\infty}^+ + \Delta \bxone_i^+ 
\hspace*{-0.5mm}
+ 
\hspace*{-0.5mm}
\int_{0^+}^{\Tones_{\infty} + \Delta \Tones_i} 
%\hspace*{-6mm}
{\rme}^{-\bAs_1 \tau} \bb_1 \ {\rm d} \tau
\right)
\label{eq3.06}
\end{eqnarray}
We now define
\begin{equation}
\bEta := 
\int_{0^+}^{\Tones_{\infty}} 
{\rm e}^{-\bAs_1 \tau} \bb_1 \ {\rm d} \tau,
\label{eq3.07}
\end{equation}
and consider the following approximate linearization of ${\rme}^{-{\bAs}_{1} \mbox{\scriptsize{$\tau$}}}$ around $\tau = \Tone_{\infty}$.
\begin{eqnarray}
\!\!\!\!\!\!
\mbox{\normalsize{$\rm e$}}^{-\bAs_1 \mbox{\scriptsize{$\tau$}}} &\approx&
{\rm e}^{{-\bAs_1} \Tones_\infty} 
-\bA_1 {\rme}^{-\bAs_1 \Tones_\infty} \left( \tau - \Tone_\infty \right) \nonumber \\
\!\!\!\!\!\!
&=& {\rm e}^{-\bAs_1 \Tones_\infty} 
\left(
\bI_4 + \bA_1 \Tone_\infty
\right)
-\bA_1 {\rme}^{-\bAs_1 \Tones_\infty} \mbox{$\tau$}
\label{eq3.08}
\end{eqnarray}
Using this equation enables the approximate calculation of the following definite integral.
\begin{eqnarray}
\mbox{\normalsize{$\displaystyle
\int_{\Tones_{\infty}}^{\Tones_{\infty} + \Delta \Tones_i} 
$}}
\mbox{\normalsize{$\rm e$}}^{-\bAs_1 \mbox{\scriptsize{$\tau$}}} 
\mbox{\normalsize{$
{\rm d} \tau 
$}}
&\approx& 
{\rm e}^{-\bAs_1 \Tones_\infty} 
\left(
\bI_4 + \bA_1 \Tone_\infty
\right) \Delta \Tone_{i} \nonumber \\
& & 
- \left(
\bA_1 {\rm e}^{-\bAs_1 \Tones_\infty} 
\right) \Tone_\infty \Delta \Tone_i \nonumber \\
&=& {\rm e}^{-\bAs_1 \Tones_\infty} \Delta \Tone_i
\label{eq3.09}
\end{eqnarray}
Following Eqs. (\ref{eq3.07}) and (\ref{eq3.09}), Eq. (\ref{eq3.06}) becomes
\begin{eqnarray}
\hspace*{-5mm}
\bxtwo_{i}^- & \approx &
\mbox{$
\left( \bI_4 + \bA_1 \mbox{$\Delta$} \Tone_i \right)
$}
{\rme}^{\bAs_1 \Tones_{\infty}} \nonumber \\
\hspace*{-5mm}
& & \times
\left(
\bxone_{\infty}^+ + \Delta \bxone_i^+ + \bEta + 
{\rme}^{-\bAs_1 \Tones_{\infty}} \bb_1 \Delta \Tone_i
\right).
\label{eq3.10}
\end{eqnarray}
Considering the relation $
\bxtwo_{\infty}^- = {\rme}^{\bAs_1 \Tones_{\infty}} \left(
\bxone_{\infty}^+ + \bEta
\right)
$ and the error terms of second order or higher as zero, Eq. (\ref{eq3.10}) can be further rearranged as follows.
\begin{eqnarray}
\bxtwo_{i}^- &=&
\left( \bI_4 + \bA_1 \Delta \Tone_i \right)
\left(
\bxtwo_{\infty}^- + {\rm e}^{\bAs_1 \Tones_{\infty}} \Delta \bxone_i^+ + \bb_1 \Delta \Tone_i
\right) \nonumber \\
& \approx &
\bxtwo_{\infty}^- + {\rm e}^{\bAs_1 \Tones_{\infty}} \Delta \bxone_i^+ + 
\left(
\bA_1 \bxtwo_{\infty}^- + \bb_1
\right)
\Delta \Tone_i
\label{eq3.11}
\end{eqnarray}
Defining the projection vector $\bp_1 := \left[ \begin{array}{cccc}
					1 & -1 & 0 & 0
					      \end{array}
\right]^{\rm T}
$, the relation $\bp_1^{\rm T} \bxtwo_{i}^- = \bp_1^{\rm T} \bxtwo_{\infty}^- = \alpha - \pi
$ holds, so multiplying both sides of Eq. (\ref{eq3.11}) by $\bp_1^{\rm T}$ from the left yields the following equation.
\begin{equation}
\bp_1^{\rm T} {\rm e}^{\bAs_1 \Tones_{\infty}} \Delta \bxone_i^+ + \bp_1^{\rm T}
\left(
\bA_1 \bxtwo_{\infty}^- + \bb _1
\right)
\Delta \Tone_i = 0
\label{eq3.12}
\end{equation}
We can solve this for $\Delta \Tone_i$ as
\begin{equation}
\Delta \Tone_i = - \frac{
\bp_1^{\rm T} {\rm e}^{\bAs_1 \Tones_{\infty}} \Delta \bxone_i^+
}{
\bp_1^{\rm T}
\left(
\bA_1 \bxtwo_{\infty}^- + \bb_1
\right)
}.
\label{eq3.13}
\end{equation}
By substituting this into Eq. (\ref{eq3.11}) and considering the relation $\bxtwo_{i}^- = \bxtwo_{\infty}^- + \Delta \bxtwo_{i}^-$, we can finally obtain
\begin{equation}
\hspace*{-1mm}
\Delta \bxtwo_{i}^- = 
\left(
\bI_4 - 
\frac{
\left(
\bA_1 \bxtwo_{\infty}^- + \bb_1
\right)
\bp_1^{\rm T}
}{
\bp_1^{\rm T}
\left(
\bA_1 \bxtwo_{\infty}^- + \bb_1
\right)
}
\right)
{\rm e}^{\bAs_1 \Tones_{\infty}} \Delta \bxone_i^+.
\label{eq3.14}
\end{equation}
The expression within the large parentheses in Eq. (\ref{eq3.14}) is defined as $\bX_1$ as follows, which can be expressed in two ways by considering the relation ${}^2 \hspace*{-0.3mm} \dot{\bx}_{\infty}^- = \bA_1 \bxtwo_{\infty}^- + \bb_1$.
\begin{equation}
\bX_1 := 
\bI_4 - 
\frac{
\left(
\bA_1 \bxtwo_{\infty}^- + \bb_1
\right)
\bp_1^{\rm T}
}{
\bp_1^{\rm T}
\left(
\bA_1 \bxtwo_{\infty}^- + \bb_1
\right)
} = 
\bI_4 - 
\frac{
{}^2
\hspace*{-0.3mm}
\dot{\bx}_{\infty}^-
\bp_1^{\rm T}
}{
\bp_1^{\rm T}
{}^2
\hspace*{-0.3mm}
\dot{\bx}_{\infty}^-
}
\label{eq3.15}
\end{equation}
Using this, the state error transition function matrix for the stance phase can be finally obtained as $\mbox{$\bQ$}_{1} := \bX_1 \mbox{\normalsize{$\rm e$}}^{\bAs_1 \Tones_{\infty}}$.

The numerical solution of $\bxtwo_{i}^-$ can be obtained as follows. To avoid the issue of $\bA_1$ being an irregular matrix, the equations of motion for the stance-leg only is formulated as follows.
\begin{equation}
\frac{\rm d}{{\rm d}t}
\left[
\begin{array}{c}
\theta_1 \\
\dth_1
\end{array}
\right] = 
\left[
\begin{array}{cc}
0 & 1 \\
\frac{2mgL}{2mL^2+I} & 0
\end{array}
\right]
\left[
\begin{array}{c}
\theta_1 \\
\dth_1
\end{array}
\right] +
\left[
\begin{array}{c}
0 \\
\frac{u_0}{2mL^2+I}
\end{array}
\right] 
\label{eq3.16}
\end{equation}
This is denoted as $\dot{\bar{\bx}} = \bar{\bA}_1 \bar{\bx} + \bar{\bb}_1$, and ${}^2\bar{\bx}_{i}^- = 
\left[
\begin{array}{cc}
{}^{2}\hspace*{-0.3mm}\theta_{1i}^- & {}^{2}\hspace*{-0.3mm}\dth_{1i}^-
\end{array}
\right]^{\rm T}
$ is then obtained numerically and instantaneously as
\begin{equation}
{}^2\hspace*{-0.3mm}\bar{\bx}_{i}^- = 
\bar{\bx} \left(
\Tone_i^-
\right) = {\rm e}^{\bar{\bAs}_1 \Tones_i} \left(
{}^1\hspace*{-0.3mm}\bar{\bx}_{i}^+ + 
\bar{\bA}_1^{-1} \bar{\bb}_1
\right) - \bar{\bA}_1^{-1} \bar{\bb}_1.
\label{eq3.17}
\end{equation}
Since the equation of motion of the swing leg is $I \ddth_2 = -u_0$, integrating this with respect to time yields the analytical solutions for the angular position and velocity as follows.
\begin{equation}
{}^2\hspace*{-0.3mm}\theta_{2i}^- = {}^1\hspace*{-0.3mm}\theta_{2i}^+ + {}^1\hspace*{-0.3mm}\dth_{2i}^+ \Tone_{1i}
- \frac{u_0 \Tone_{1i}^2}{2I}, \ \ 
%\label{eq3.14a}
{}^2\hspace*{-0.3mm}\dth_{2i}^- = {}^1\hspace*{-0.3mm}\dth_{2i}^+ - \frac{u_0 \Tone_{1i}}{I}
\label{eq3.18}
\end{equation}
Following Eqs. (\ref{eq3.17}) and (\ref{eq3.18}), the $\Tone_i$ satisfying $\bp_1^{\rm T} \bxtwo_{i}^- = \alpha - \pi$ can be found instantaneously using the bisection method.

\subsection{Collision Phase 2}

Since no angular position update occurs during this collision phase, considering the angular velocity update rule determined by Eq. (\ref{eq2.09}) defines the transition equation as
\begin{equation}
\bxtwo_i^+ = \bR_2 \bxtwo_i^-,
\label{eq3.19}
\end{equation}
where
\begin{eqnarray}
\bR_2 &:=& 
\left[
\begin{array}{cc}
\bI_2 & {\bf 0}_{2 \times 2} \\
{\bf 0}_{2 \times 2} & 
\bI_2 - \bM^{-1} \bS \left(
\bS^{\rm T} \bM^{-1} \bS
\right)^{-1} \bS^{\rm T}
\end{array}
\right] \nonumber \\
&=&
\left[
\begin{array}{cccc}
1 & 0 & 0 & 0 \\
0 & 1 & 0 & 0 \\
0 & 0 & 1-\frac{I}{2(mL^2+I)} & \frac{I}{2(mL^2+I)} \\ 
0 & 0 & 1-\frac{I}{2(mL^2+I)} & \frac{I}{2(mL^2+I)}
\end{array}
\right].
\label{eq3.20}
\end{eqnarray}
In the steady-state motion, Eq. (\ref{eq3.19}) is described as
\begin{equation}
\bxtwo_\infty^+ = \bR_2 \bxtwo_\infty^-.
\label{eq3.21}
\end{equation}
Subtracting Eq. (\ref{eq3.21}) from Eq. (\ref{eq3.19}), we obtain $\Delta \bxtwo_i^+ = \bR_2 \Delta \bxtwo_i^-$, which indicates that the state error transition function matrix for the collision phase is $\bR_2$. The four eigenvalues of $\bR_2$ are a triple root at $1$ and $0$, and CP2 is a stable phase.

\subsection{Swing Phase 2}

The time variable $t$ is reset to zero again at CP2, and the collision time when the swing leg collides with the ground surface is defined as $t = \Ttwo_i$. By considering the linear approximation of Eq. (\ref{eq3.03}) for the constrained dynamics of Eq. (\ref{eq2.14}), the constrained linearized equation of motion becomes
\begin{equation}
\dot{\bx} = \bA_2 \bx,
\label{eq3.22}
\end{equation}
where
\begin{eqnarray}
\bA_2 &=&
\left[
\begin{array}{cc}
{\bf 0}_{2 \times 2} & \bI_2 \\
\bM^{-1} \!\!
\left(
\bS \left( \bS^{\rm T} \bM^{-1} \! \bS \right)^{-1} \!\!\! \bS^{\rm T} \bM^{-1} \! - \! \bI_2
\right)
\bG_0 & {\bf 0}_{2 \times 2}
\end{array}
\right] \nonumber \\
&=& 
\left[
\begin{array}{cccc}
0 & 0 & 1 & 0 \\
0 & 0 & 0 & 1 \\
\frac{m g L}{mL^2+I} & 0 & 0 & 0 \\
\frac{m g L}{mL^2+I} & 0 & 0 & 0
\end{array}
\right].
\label{eq3.23}
\end{eqnarray}
Considering the relation $\bxone_{\infty}^- = {\rme}^{\bAs_2 \Ttwos_{\infty}} \bxtwo_{\infty}^+$ and the error terms of second order or higher as zero, $\bxone_{i+1}^-$ can be arranged and approximated as follows.
\begin{eqnarray}
\bxone_{i+1}^- &=& \bx \left( \Tone_i^- \right) = {\rm e}^{\bAs_2 \Ttwos_i} \bxtwo_i^+ 
\label{eq3.24}
\\
&=& {\rme}^{\bAs_2 \Ttwos_{\infty}} {\rme}^{\bAs_2 \Delta \Ttwos_i}
\left(
\bxtwo_{\infty}^+ + \Delta \bxtwo_i^+ 
\right) 
\nonumber \\
%\hspace*{-5mm}
%\bxone_{i+1}^- 
& \approx &
{\rme}^{\bAs_2 \Ttwos_{\infty}} 
\mbox{$
\left( \bI_4 + \bA_2 \mbox{$\Delta$} \Ttwo_i \right)
$} \left(
\bxtwo_{\infty}^+ + \Delta \bxtwo_i^+ 
\right)
\nonumber \\
%\hspace*{-5mm}
&\approx&
\bxone_{\infty}^- + {\rme}^{\bAs_2 \Ttwos_{\infty}} \Delta \bxtwo_i^+ + 
%{\rme}^{\bAs_2 \Ttwos_{\infty}} 
\bA_2 \bxone_{\infty}^-
\Delta \Ttwo_i
\label{eq3.25}
\end{eqnarray}
Defining the projection vector $\bp_2 := \left[ \begin{array}{cccc}
					1 & 0 & 0 & 0
					      \end{array}
\right]^{\rm T}
$, the relation $\bp_2^{\rm T} \bxone_{i+1}^- = \bp_2^{\rm T} \bxone_{\infty}^- = \frac{\alpha}{2} + \phi$ holds, so multiplying both sides of Eq. (\ref{eq3.25}) by $\bp_2^{\rm T}$ from the left yields the following equation.
\begin{equation}
\bp_2^{\rm T} {\rme}^{\bAs_2 \Ttwos_{\infty}} \Delta \bxtwo_i^+ + 
\bp_2^{\rm T} \bA_2 \bxone_{\infty}^-
\Delta \Ttwo_i = 0
\label{eq3.26}
\end{equation}
We can solve this for $\Delta \Ttwo_i$ as
\begin{equation}
\Delta \Ttwo_i = - \frac{
\bp_2^{\rm T} {\rme}^{\bAs_2 \Ttwos_{\infty}} \Delta \bxtwo_i^+
}{
\bp_2^{\rm T} \bA_2 \bxone_{\infty}^-
}.
\label{eq3.27}
\end{equation}
By substituting this into Eq. (\ref{eq3.25}) and considering the relation $\bxone_{i+1}^- = \bxone_{\infty}^- + \Delta \bxone_{i+1}^-$, we can finally obtain
\begin{equation}
\Delta \bxone_{i+1}^- = 
\left(
\bI_4 - 
\frac{
\bA_2 \bxone_{\infty}^-
\bp_2^{\rm T}
}{
\bp_2^{\rm T}
\bA_2 \bxone_{\infty}^-
}
\right)
{\rm e}^{\bAs_2 \Ttwos_{\infty}} \Delta \bxtwo_i^+.
\label{eq3.28}
\end{equation}
The expression within the large parentheses in Eq. (\ref{eq3.28}) is defined as $\bX_2$ as follows, which can be expressed in two ways by considering the relation ${}^1\dot{\bx}_{\infty}^- = \bA_2 \bxone_{\infty}^-$.
\begin{equation}
\bX_2 := 
\bI_4 - 
\frac{
\bA_2 \bxone_{\infty}^-
\bp_2^{\rm T}
}{
\bp_2^{\rm T}
\bA_2 \bxone_{\infty}^-
} = 
\bI_4 - 
\frac{
{}^1
\hspace*{-0.3mm}
\dot{\bx}_{\infty}^-
\bp_2^{\rm T}
}{
\bp_2^{\rm T}
{}^1
\hspace*{-0.3mm}
\dot{\bx}_{\infty}^-
}
\label{eq3.29}
\end{equation}
Using this, the state error transition function matrix for the stance phase can be finally obtained as $\mbox{$\bQ$}_{2} := \bX_2 \mbox{\normalsize{$\rm e$}}^{\bAs_2 \Ttwos_{\infty}}$.

The numerical solution for ${}^{1} \hspace*{-0.3mm} \bar{\bx}_{i+1}^-$ can be obtained instantaneously by using the bisection method to find the $\Ttwo_i$ for which the first component of Eq. (\ref{eq3.24}) equals $\phi + \frac{\alpha}{2}$. Then, by iteratively computing the periods of the swing phases and the states at impact numerically and discretely, calculations for approximately 1000 steps of walking can be completed with zero perceived time.
\section{GAIT ANALYSIS}

\subsection{Poincar\'{e} Map and Dimension Reduction}

Based on the discussion in the previous section, the state error transition from immediately before the $i$-th CP1 to immediately before the $i+1$-th CP1 is described by
\begin{equation}
\Delta \bxone_{i+1}^- = 
\bQ_2 \bR_2 \bQ_1 \bR_1 \Delta \bxone_i^-.
\label{eq4.01}
\end{equation}
Since the robot walks while achieving constraint on impact posture, the angular errors of both legs during CP1 are always zero and the angular velocity errors of those are the same. Therefore, the state error vector immediately before CP1 is
\begin{equation}
\Delta \bxone_{i}^{-} = 
\left[
\begin{array}{cccc}
0 & 0 & \Delta {}^{1}\hspace{-0.3mm}\dth_{1i}^- & \Delta {}^{1}\hspace{-0.3mm}\dth_{1i}^-
\end{array}
\right]^{\rm T}.
\label{eq4.02}
\end{equation}
Utilizing this property, the state error transition function matrix can be reduced to a scalar. To simplify notation, the angular velocities of both legs immediately before the $i$-th CP1 are denoted as $\omega_i := {}^{1}\hspace*{-0.3mm}\dth_{1i}^- = {}^{1}\hspace*{-0.3mm}\dth_{2i}^-$ below.

%The state error transition function matrix from immediately before the $i$-th CH1 to the next becomes $\bP := \bQ_2 \bR_2 \bQ_1 \bR_1$.
%$\omega := \dth_1^- = \dth_2^-$

The following relations hold between it and the original state error vector.
\begin{equation}
\!\!
\Delta \bxone_{i}^{-} = 
\left[
\begin{array}{cccc}
0 & 0 & 1 & 1
\end{array}
\right]^{\rm T} \Delta \omega_i, \ 
\Delta \omega_{i+1} =\left[
\begin{array}{cccc}
0 & 0 & 1 & 0
\end{array}
\right] \Delta \bxone_{i+1}^{-}
\label{eq4.03}
\end{equation}
The state error transition function of Eq. (\ref{eq4.01}) is then reduced to $\Delta \omega_{i+1} = F \Delta \omega_{i}$ where
\begin{equation}
F := 
\left[
\begin{array}{cccc}
0 & 0 & 1 & 0
\end{array}
\right] \bQ_2 \bR_2 \bQ_1 \bR_1
\left[
\begin{array}{cccc}
0 & 0 & 1 & 1
\end{array}
\right]^{\rm T}.
\label{eq4.04}
\end{equation}
While symbolic calculation for $F$ is possible using Mathematica \cite{Mathematica}, the formula is highly complex, so its details are omitted. In this section, rather than using the specific formula of $F$, we will compute it numerically according to Eq. (\ref{eq4.04}). If the absolute value of $F$ is less than 1, the generated wheel is determined to be asymptotically stable.

\begin{figure}[!b]
\centering
\footnotesize
\vspace*{-2mm}
\hspace*{0.5mm}
\scalebox{0.66}{\input{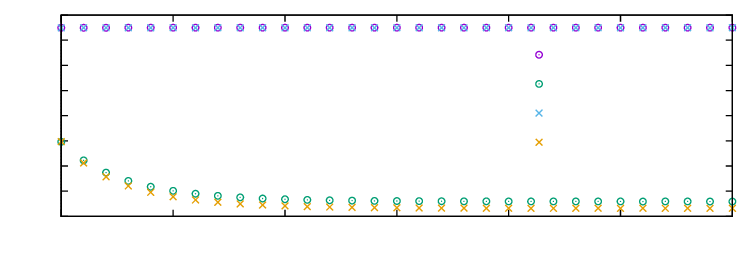}}\\
\vspace*{-1mm}
(a) Step periods, $\Tone_i$ and $\Ttwo_i$\\
\hspace*{0.5mm}
\scalebox{0.66}{\input{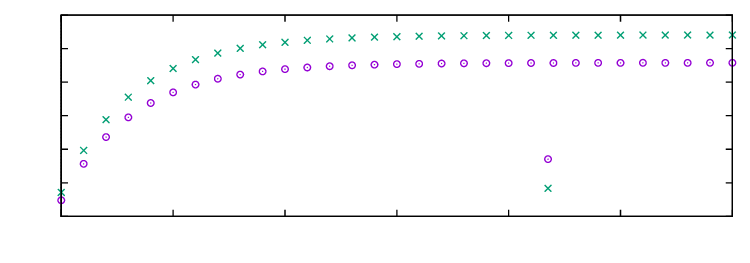}}\\
\vspace*{-1mm}
(b) Angular velocities immediately before CP1
\vspace*{-2mm}
\caption{Evolution of gait descriptors in generated wheel gait of nonlinear and linearized models}
\label{fig4.01}
\end{figure}

\begin{figure}[!b]
%\vspace*{0.5mm}
\centering
\footnotesize
%\vspace*{0.5mm}
\hspace*{0.5mm}
\scalebox{0.66}{\input{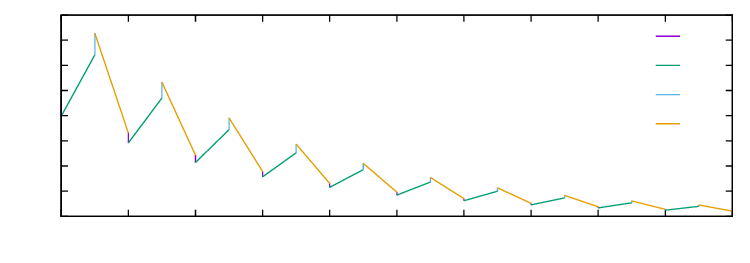}}\\
\vspace*{-2mm}
\caption{Evolution of state error norm in generated wheel gait for linearized models of Fig. \ref{fig4.01}}
\label{fig4.02}
%\vspace*{-3mm}
\end{figure}

\subsection{Typical Wheel Gait}

Figure \ref{fig4.01} shows the evolution of the gait descriptors in the generated wheel gait of the nonlinear and linearized models. Here, (a) is the step periods and (b) is the angular velocity immediately before CP1. The system parameters and initial condition were set to the same values as in the wheel gait of Fig. \ref{fig2.03}. From Fig. \ref{fig4.01}(a), we see that $\Tone_i$ is almost identical for both models, while $\Ttwo_i$ is slightly smaller for the linearized model. Figure \ref{fig4.01}(b) shows that the entire angular velocity immediately before CP1 converges to a clearly larger value in the linearized model. This is likely because the linearized model yields greater restored mechanical energy due to gravity. While detailed discussion is omitted here, we recently developed a method to obtain a more accurate linearized model by generalizing the expansion point in the approximate linearization of the gravity term. We plan to use this method in the future to report improved analytical results.

Figure \ref{fig4.02} shows the time evolution of the state error norm in the generated gait for the linearized model of Fig. \ref{fig4.01}. Note that for clarity, if the timing of CP1 is at the $i$-th impact, the timing of the subsequent CP2 is plotted at the $i+0.5$-th impact. The state error norm increases in SP1 and CP2 and decreases in SP2 and CP1. This indicates that SP1 and CP2 are unstable, while SP2 and CP1 are stable. The gait as a whole is asymptotically stable, and it can be seen that the value of $F$ defined by Eq. (\ref{eq4.03}) will be greater than zero and less than one. It can be said that the high stability of SP2 contributes most significantly to achieving this asymptotic stability. Note that the value of $F$ in this wheel gait was calculated through iterative computation to be 0.734577.

\subsection{Effect of $u_0$}

\begin{figure}[!b]
\vspace*{-2mm}
\begin{minipage}[t]{0.49\columnwidth}
\centering
\hspace*{-5mm}
\scalebox{0.46}{
\input{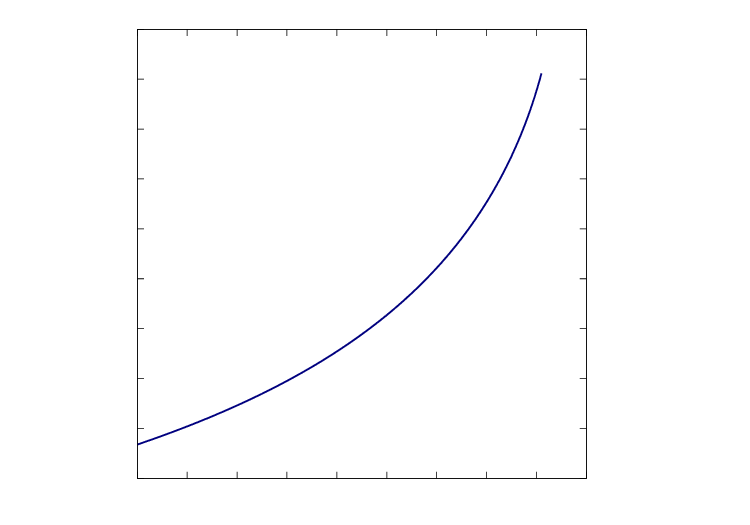}
}
\end{minipage}
%\hspace*{1mm}
\begin{minipage}[t]{0.49\columnwidth}
\centering
\hspace*{-5mm}
\scalebox{0.46}{
\input{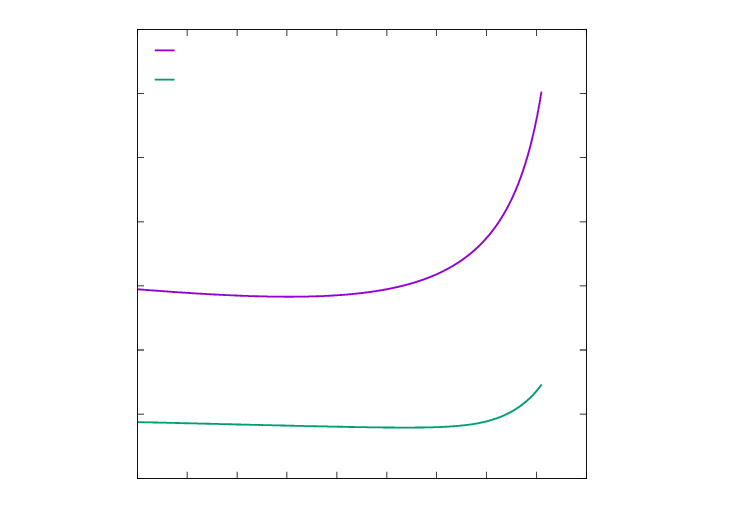}
}
\end{minipage}
\vspace*{3mm}

\begin{minipage}[t]{0.49\columnwidth}
\centering
\hspace*{-5mm}
\scalebox{0.46}{
\input{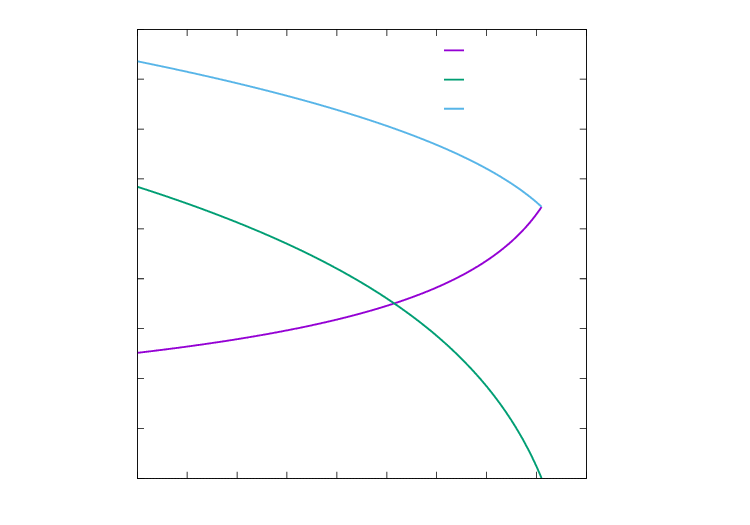}
}
\end{minipage}
%\hspace*{1mm}
\begin{minipage}[t]{0.49\columnwidth}
\centering
\hspace*{-5mm}
\scalebox{0.46}{
\input{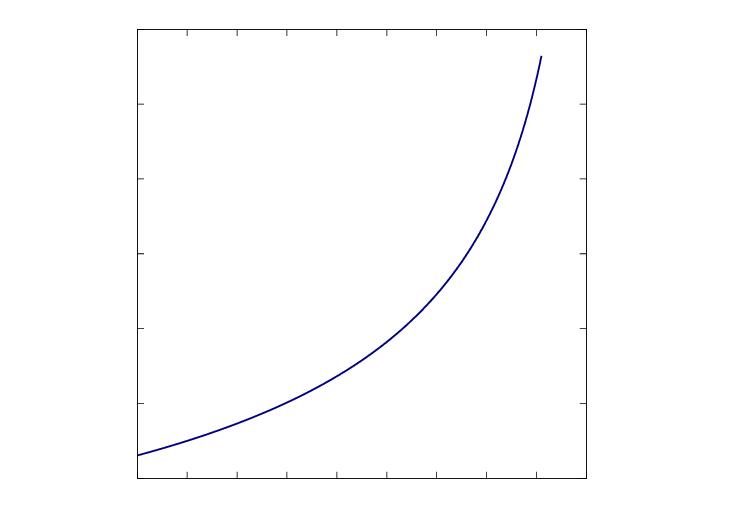}
}
\end{minipage}
\vspace*{1mm}
\caption{Gait descriptors with respect to $u_0$ in linearized model}
\label{fig4.03}
%\vspace*{-3mm}
\end{figure}

Figure \ref{fig4.03} shows the gait descriptors of the linearized model with respect to the constant hip-joint torque, $u_0$. The system parameters other than $u_0$ were set to the same values as in Table \ref{table01} again. Here, (a) is the $F$, (b) the maximum singular values of $\bQ_1$ and $\bQ_2$, (c) the steady step periods of SP1 and SP2, and (d) the steady angular velocity immediately before CP1, $\omega_{\infty}$. First, we set $u_0$ to $-10$ [N$\cdot$m], performed $1000$ steps of walking using iterative calculations on the linearized model, saved the data for the last 20 steps, and calculated its average. Then, we incremented $u_0$ by $0.01$ [N$\cdot$m] and repeated this process as long as stable gait generation remained possible. Since this calculation method does not use numerical integration, all computations were completed in about one minute.

Fig. \ref{fig4.03}(a) shows that the value of $F$ monotonically increases with increasing $u_0$, indicating a deterioration in gait convergence. Despite the target impact posture (energy-loss coefficient) and the restored mechanical energies from gravity and hip-joint torque being constant, the convergence speed changes. This is thought to occur because the kinetic energy lost due to the mechanical stopper varies with $u_0$, as discussed later.

Fig. \ref{fig4.03}(b) shows that both values of $\bar{\sigma} \left( \bQ_1 \right)$ and $\bar{\sigma} \left( \bQ_2 \right)$ decrease monotonically for a while as $u_0$ increases, but they rise sharply after passing through their minimum values. As seen in the results of Fig. \ref{fig4.03}(a), as $u_0$ approaches the upper limit, the gait convergence speed significantly slows down. As predicted by the results in Fig. \ref{fig4.02}, the values of $\bar{\sigma} \left( \bQ_1 \right)$ are substantially larger than those of $\bar{\sigma} \left( \bQ_2 \right)$ across the entire range, reaffirming the reason for the significant increase in the state error norm in SP1. Although the values of $\bar{\sigma} \left( \bQ_2 \right)$ consistently exceeds $1.0$, the reduction in the state error norm at SP2 is guaranteed. This is likely because the collision at the target impact posture at the end of SP2 resets the angular error norm to zero. Further detailed investigation is needed regarding the reduction in the angular velocity error norm. Since both $\bR_1$ and $\bR_2$ are constant matrices with maximum singular values that remain constant regardless of $u_0$, the plots have been omitted. Calculations for $\bR_1$ yielded $\bar{\sigma} \left( \bR_1 \right) = 1.18262$, while the remaining three were 1, 1, and 0.0939537. The two singular values being 1 is because the angular error does not change in this phase. Since the angular velocity error transitions in a direction that reduces in this phase, it is stable. Calculations for $\bR_2$ yielded $\bar{\sigma} \left( \bR_2 \right) = 1.28062$, while the remaining three were 1, 1, and 0. The reason the two singular values are 1 is the same as before. Since the angular velocity error transitions in a direction that increases in this phase, it is unstable.

Fig. \ref{fig4.03}(c) shows that $\Tone_{\infty}$ increases monotonically with increasing $u_0$, while both $\Ttwo_{\infty}$ and $\Tone_{\infty} + \Ttwo_{\infty}$ decrease monotonically. As $u_0$ becomes larger -- in this case, approaching zero -- the upper limit of $u_0$ is the value reached when $\Ttwo_{\infty}$ decreases to zero. These results indicate that as the driving force weakens, the period until collision with the mechanical stoppers lengthens, while simultaneously, the period from collision with the stoppers until the next collision with the ground shortens. It can be concluded that, even with weak driving torque, maintaining drive throughout the entire single-limb support phase contributes to generating high-speed walking motion. This is supported by the fact that the steady angular velocity immediately before CP1 monotonically increases with increasing $u_0$, as shown in the results of Fig. \ref{fig4.03}(d).

\subsection{Discussion}

As discussed in Section II-F, in the generated gait, the robot's total mechanical energy increased monotonically during SP1 and significantly and discontinuously decreased during CP2 in each step. We now define the kinetic energy lost in CP2 as
\begin{equation}
\Delta {}^2 \! K_i := {}^2 \! K_i^- - {}^2 \! K_i^+ > 0.
\label{eq4.05}
\end{equation}
We also define $\Delta E_g$ and $\Delta E_u$ as the restored mechanical energies due to gravity and control torque, respectively. These are specifically determined by
\begin{eqnarray*}
\Delta E_g &=& 4 m g L \sin \frac{\alpha}{2} \sin \phi, \\
\Delta E_u &=& \int_{0^+}^{\Tones_i^-} \! \dot{\bth}^{\rm T} \bS u \ {\rm d}\tau = 
\int_{-\alpha}^{\alpha - \pi} \!\! u_0 \ {\rm d} \! \left( \theta_1 - \theta_2 \right) = - \frac{\pi}{2} u_0.
\label{eq4.06}
\end{eqnarray*}
Since both are positive constants, the total restored mechanical energy $\Delta E := \Delta E_g + \Delta E_u$ for one step also remains positive constant. 

From the above, the following recurrence formula holds for the kinetic energy immediately before CP1.
\begin{equation}
{}^1 \! K_{i+1}^- = \varepsilon {}^1 \! K_{i}^- + \Delta E - \Delta K_i
\label{eq4.07}
\end{equation}
Here, $\varepsilon := {}^1 \! K_{i}^- / {}^1 \! K_{i}^+$ is the energy-loss coefficient and is a positive constant less than $1$. If $\Delta K_i = 0$, this recurrence formula becomes identical to that of a conventional RW, immediately proving that the generated gait is period-1 and asymptotically stable. In addition, since $\varepsilon$ is constant in this case, the convergence speed remains constant regardless of $\Delta E$. It is precisely because $\Delta K_i$ is nonzero that the calculation the complex state error transition function matrices described in the previous section becomes necessary. If we could demonstrate the convergence of this $\Delta K_i$, the asymptotic stability of the generated wheel gait could be explained more clearly from the perspective of mechanical energy balance. Furthermore, since $\Delta K_i$ changes complexly depending on $u_0$, it can be considered that the convergence speed also changes, as seen in the value of $F$ in Fig. \ref{fig4.02}(a).

\section{CONCLUSION AND FUTURE WORK}

This paper discusses the self-stabilization principle inherent in wheel gaits with constraint on impact posture for a planar X-shaped walker composed of two identical leg frames with mechanical stoppers, using the state error transition function matrices. Analysis of the characteristics of the four phases constituting one step of asymptotically stable walking revealed that SP1 and CP2 are unstable, while SP2 and CP1 are stable. While the high stabilizing effect of SP2 undoubtedly contributes significantly to the overall stability of the wheel gait, the precise manner in which the phases influence each other remains poorly understood. Further theoretical investigations are required.

The gait generation method and mechanism proposed in this paper is extremely simple, requiring only driving the hip joint with a constant torque and not needing real-time feedback information. As long as the collision timing for support-leg exchange can be detected, stable wheel gait can be generated simply by continuously applying a constant torque to the hip joint. This can be considered a highly advantageous method for actual hardware implementation. To generate stable wheel gait on a horizontal plane, it is necessary to add a reaction wheel between the leg frames, doubling the number of driving torques \cite{Kiefer,Spong}. Furthermore, discussions regarding the stability of the reaction wheels' motion, which behave as zero dynamics, are also required. This will be addressed as a future research topic.

%\addtolength{\textheight}{-12cm}

\end{document}